\documentclass[runningheads]{llncs}
\usepackage[T1]{fontenc}
\usepackage{float}
\usepackage{graphicx}
\usepackage{booktabs}
\usepackage{multirow}
\usepackage{amsmath}
\usepackage{amssymb}

\begin{document}
\title{Conformal Prediction for Risk-Controlled Medical Entity Extraction Across Clinical Domains}
\titlerunning{Conformal Prediction}
%
\author{Manil Shrestha \and
Edward Kim}
\authorrunning{M. Shrestha and E. Kim}
%
\institute{
Department of Computer Science, Drexel University, Philadelphia, USA\\
\email{\{ms5267, ek826\}@drexel.edu}}
\maketitle              
\begin{abstract}
Large Language Models (LLMs) are increasingly used for medical entity extraction, yet their confidence scores are often miscalibrated, limiting safe deployment in clinical settings. We present a conformal prediction framework based on risk-controlling prediction sets~\cite{bates2021distribution} that provides finite-sample false discovery rate (FDR) guarantees for LLM-based extraction across two clinical domains. First, we extract structured entities from 1,000 FDA drug labels across eight sections using GPT-4.1, verified via FactScore-based atomic statement evaluation (97.7\% accuracy over 128,906 entities). Second, we extract radiological entities from MIMIC-CXR reports using the RadGraph schema with GPT-4.1 and Llama-4-Maverick, evaluated against physician annotations (entity F1: 0.83--0.84). Our central finding is that miscalibration direction reverses across domains: on well-structured FDA labels, models are underconfident, and the global baseline FDR of 2.3\% trivially satisfies $\alpha = 0.05$, though per-section analysis reveals that three sections require 41--100\% rejection. On free-text radiology reports, models are overconfident, and FDR control at $\alpha = 0.10$ produces sharply different outcomes across models: Llama-4-Maverick rejects 19.6\% of extractions while GPT-4.1 rejects 59.3\%, with both models rejecting all uncertain observations. Sweep analysis across $\alpha$ values reveals sharp transitions in acceptance behavior that expose the baseline error structure of each domain. These results demonstrate that calibration is not a global model property but depends on document structure, extraction category, and model architecture, motivating domain-specific conformal calibration for safe clinical deployment.

\end{abstract}
\section{Introduction}
Extracting structured entities from unstructured medical text is critical for drug safety surveillance, clinical coding, and biomedical research. Large Language Models (LLMs) have shown strong performance on these tasks~\cite{kim2024structured,wu2024framework}, achieving near-expert accuracy on entity extraction from electronic health records (EHRs), drug labels, and radiology reports. However, deploying LLMs in clinical settings requires not only high accuracy but also reliable uncertainty quantification: a model that assigns high confidence to incorrect extractions can lead to silent errors propagating through clinical decision pipelines.

LLM softmax probabilities are poorly calibrated in general~\cite{geng2024survey,guo2017calibration}. Models can be systematically overconfident, assigning near-certain probabilities to incorrect predictions, or underconfident, assigning low probabilities to correct ones. Standard post-hoc calibration techniques such as temperature scaling~\cite{guo2017calibration} require held-out validation data and provide no formal coverage guarantees. Moreover, calibration quality varies significantly across tasks, domains, and entity types, making it difficult to establish a single reliability threshold.

Conformal prediction (CP) addresses these challenges by converting raw model scores into accept/reject decisions with finite-sample guarantees under only an exchangeability assumption~\cite{angelopoulos2023conformal}. Bates et al.~\cite{bates2021distribution} extended this framework to risk-controlling prediction sets, which provide guarantees on the expected loss under arbitrary monotone loss functions, including false discovery rate (FDR) control. Rather than requiring well-calibrated probabilities, CP adapts its acceptance threshold to the empirical distribution of scores, automatically becoming more conservative when models are poorly calibrated and more permissive when they are well-calibrated. This makes CP particularly appealing for clinical deployment, where regulatory requirements demand provable reliability bounds.

In this work, we apply FDR-controlling conformal prediction to LLM-based medical entity extraction across two fundamentally different clinical domains: (1) structured FDA drug labels, where we extract entities from eight standardized sections, and (2) free-text radiology reports, where we extract entities and relations following the RadGraph schema~\cite{radgraph}. These domains differ in document structure, vocabulary, and annotation methodology, allowing us to study how calibration behavior varies across clinical contexts.

Our contributions are: (1) a conformal prediction framework for medical entity extraction that provides finite-sample FDR guarantees across heterogeneous clinical domains, ensuring that the proportion of accepted but incorrect extractions is bounded; (2) the empirical finding that LLM miscalibration direction reverses across domains, with models underconfident on structured FDA labels and overconfident on free-text radiology reports; (3) sweep analysis across miscoverage rates revealing sharp transitions in acceptance behavior that expose the baseline error structure of each domain and the critical difference between global and per-category thresholds; (4) cross-model comparison showing that FDR-controlling thresholds depend not only on extraction accuracy but on how well a model's confidence scores discriminate between correct and incorrect extractions.

\section{Related Work}

\paragraph{LLM-Based Medical Entity Extraction.}
LLMs have been applied to a range of clinical extraction tasks. Kim et al.~\cite{kim2024structured} demonstrated structured extraction from medical knowledge bases using GPT-4, while Wu et al.~\cite{wu2024framework,wu2025leveraging} developed frameworks for extracting adverse events from FDA drug labeling documents using LLMs with regulatory-grade evaluation. For radiology reports, the RadGraph dataset~\cite{radgraph} provides physician-annotated entities and relations from MIMIC-CXR chest X-rays~\cite{johnson2016mimic}, enabling systematic evaluation of extraction systems.

\paragraph{Confidence and Calibration in LLMs.}
Geng et al.~\cite{geng2024survey} survey confidence estimation methods for LLMs, highlighting that token-level log-probabilities, while informative, do not reliably indicate factual correctness. Guo et al.~\cite{guo2017calibration} showed that modern neural networks are systematically miscalibrated, a finding that extends to LLMs. Laban et al.~\cite{laban2023you} demonstrated that LLM confidence can be fragile under adversarial probing. These findings motivate distribution-free approaches like conformal prediction.

\paragraph{Conformal Prediction in Medicine and NLP.}
Conformal prediction provides distribution-free coverage guarantees under exchangeability~\cite{angelopoulos2023conformal}. Bates et al.~\cite{bates2021distribution} extended this to risk-controlling prediction sets, enabling control of arbitrary monotone loss functions including FDR. In medicine, CP has been applied to skin lesion classification~\cite{fayyad2024empirical}, genomic analysis~\cite{papangelou2025reliable}, medical coding~\cite{snyder2024conformal}, and EHR mining~\cite{genari2025mining}. Vazquez and Facelli~\cite{vazquez2022conformal} survey CP applications across clinical domains. Campos et al.~\cite{campos2024conformal} provide a comprehensive survey of CP for NLP tasks. In our prior work~\cite{kim2025conformal}, we applied CP to LLM-based EHR extraction across 10,000 visits, validating thresholds against expert consensus using the VeriFact framework~\cite{chung2025verifact}. The present work extends this approach to FDA labels and radiology reports, revealing domain-dependent calibration behavior not observed in the EHR setting.

\section{Methods}

We apply FDR-controlling conformal prediction to two extraction tasks using GPT-4.1~\cite{achiam2023gpt} and Llama 4 Maverick (17B parameters, 128 experts)~\cite{meta2025llama}. Both models return per-token log-probabilities, enabling span-level confidence estimation. The pipeline has four steps.

\subsection{Step 1: Extraction with Token Confidence}

\paragraph{FDA Drug Labels.}
We extract entities from 1,000 FDA drug labels across eight standardized sections: indications and usage, adverse reactions, drug interactions, contraindications, warnings, pregnancy, pediatric use, and geriatric use. Each drug label is processed by GPT-4.1, which extracts entities (drug names, conditions, adverse events, anatomical structures, etc.) along with per-token log-probabilities. The source dataset comprises 2,437 asthma-indicated drugs from the FDA's openFDA API.

\paragraph{Radiology Reports.}
We extract entities and relations from 100 MIMIC-CXR chest X-ray reports~\cite{johnson2016mimic} following the RadGraph schema~\cite{radgraph}. The schema defines four entity categories: anatomy definitely present (\textsc{anat-dp}), observation definitely present (\textsc{obs-dp}), observation definitely absent (\textsc{obs-da}), and observation uncertain (\textsc{obs-u}). Relations include \textit{located\_at}, \textit{modify}, and \textit{suggestive\_of}. We evaluate both GPT-4.1 and Llama-4-Maverick with 5-shot prompting (examples drawn from the RadGraph dev set) and zero-shot prompting.

\paragraph{Span Confidence.}
For each entity span of $m_e$ tokens with softmax probabilities $p_{e,t}$, we compute a span-level confidence via the geometric mean:
\begin{equation}
\hat{p}_e = \exp\!\left(\frac{1}{m_e}\sum_{t=1}^{m_e} \log p_{e,t}\right)
\end{equation}
This aggregation is motivated by the observation that span correctness depends on all constituent tokens: a single low-probability token (e.g., a misspelled drug name or wrong anatomical modifier) typically indicates an extraction error. The geometric mean is more sensitive to such outlier tokens than the arithmetic mean, providing a more informative confidence signal.

\subsection{Step 2: Verification}

\paragraph{FDA Labels.}
Verification uses an LLM-as-a-judge approach following the VeriFact framework~\cite{chung2025verifact}. GPT-5-mini evaluates each extracted entity against the source FDA label text, producing a fact-score on a 0--3 scale: 0 (hallucinated/unsupported), 1 (partially supported), 2 (mostly supported), and 3 (fully verified). Only score 3 counts as correct ($y_e = 1$). This strict threshold ensures that conformal calibration is performed against a high-quality ground truth. Over 128,906 entities, 97.7\% received a fact-score of 3.

\paragraph{Radiology Reports.}
For RadGraph, we evaluate against physician-annotated gold entities from the RadGraph test set (100 reports, 2,763 gold entities). An extracted entity is marked correct ($y_e = 1$) if it exactly matches a gold entity in both text span and label. This exact-match criterion is stricter than token-overlap matching but ensures clinically meaningful evaluation.

\subsection{Step 3: Nonconformity Score}

We define the nonconformity score as the logit of span confidence:
\begin{equation}
s_e = \mathrm{logit}(\hat{p}_e) = \log\!\left(\frac{\hat{p}_e}{1 - \hat{p}_e}\right)
\end{equation}
This transformation maps confidence from $(0,1)$ to $(-\infty,+\infty)$, spreading out the high-confidence region $[0.9, 1.0)$ where most entities cluster. Higher scores indicate higher confidence. The logit transformation is standard in conformal prediction for probability-valued scores~\cite{angelopoulos2023conformal}.

\subsection{Step 4: FDR-Controlling Conformal Calibration}

We partition verified entities into calibration (50\%) and test (50\%) sets using a fixed random seed. Rather than targeting marginal coverage, we calibrate thresholds to control the false discovery rate among accepted extractions, following the risk-controlling prediction sets framework of Bates et al.~\cite{bates2021distribution}.

For each domain $d$ and extraction category $c$, we select the threshold $\tau_{d,c}$ as the smallest value such that the empirical FDR on the calibration set does not exceed $\alpha$:
\begin{equation}
\tau_{d,c} = \inf\left\{t : \frac{\sum_{e \in \mathcal{D}^{\text{cal}}_{d,c}} \mathbf{1}(y_e = 0)\cdot\mathbf{1}(s_e \geq t)}{\max\!\left(1,\;\sum_{e \in \mathcal{D}^{\text{cal}}_{d,c}} \mathbf{1}(s_e \geq t)\right)} \leq \alpha\right\}
\end{equation}

Entities with $s_e \geq \tau_{d,c}$ are accepted; the rest are flagged for human review. This guarantees that the expected proportion of accepted but incorrect extractions is bounded by $\alpha$:
\begin{equation}
\mathbb{E}\left[\frac{1}{|A_{d,c}|} \sum_{e \in A_{d,c}} \mathbf{1}(y_e = 0)\right] \leq \alpha
\end{equation}
where $A_{d,c} = \{e : s_e \geq \tau_{d,c}\}$ is the accepted set.

We evaluate at $\alpha = 0.05$ for FDA labels and $\alpha = 0.10$ for RadGraph. For both domains, we additionally conduct sweep analysis across multiple $\alpha$ values to characterize the precision-rejection tradeoff. For RadGraph, we compute both global thresholds and per-category thresholds $\tau_c$, since calibration behavior varies substantially across entity types.

\section{Results and Discussion}

\subsection{FDA Label Extraction}

We extracted 128,906 entities from 1,000 FDA drug labels across eight sections using GPT-4.1, of which 110,664 have both confidence scores and fact-score verification. Overall fact-score accuracy is 97.7\% (108,107/110,664 entities scored as fully verified).

\begin{figure}[h]
  \centering
  \includegraphics[width=1\textwidth]{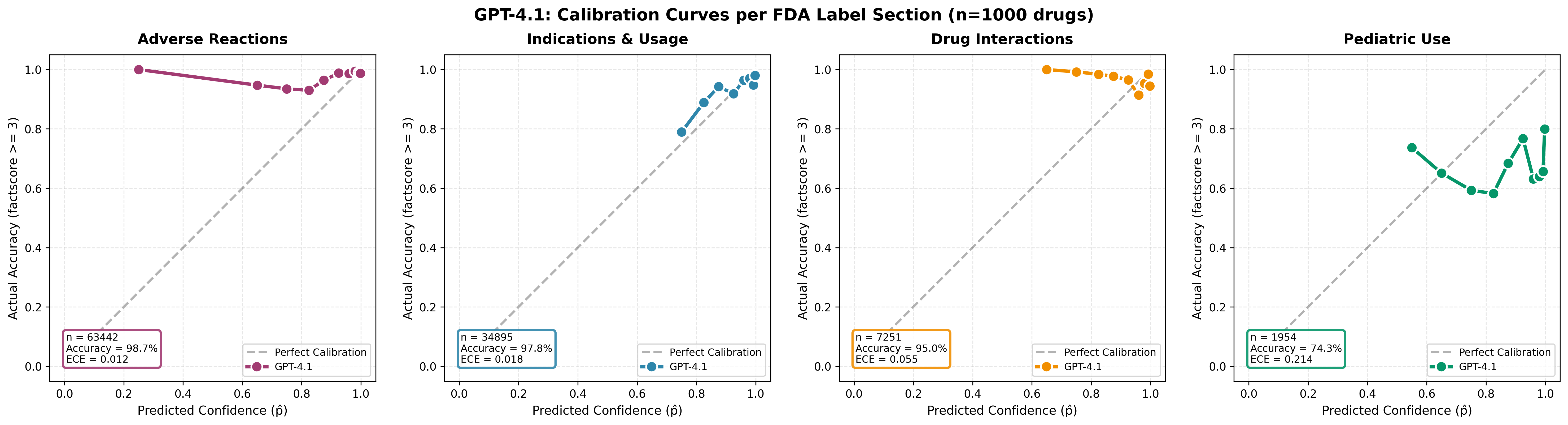}
  \caption{Calibration curves for four sections from FDA label extraction with GPT-4.1 ($n{=}1{,}000$ drugs). Predicted span confidence $\hat{p}$ (geometric mean of token probabilities) versus empirical accuracy. The model is systematically underconfident across most sections: curves lie above the diagonal. Pediatric Use is an exception, exhibiting overconfidence. ECE ranges from 0.012 (Adverse Reactions) to 0.214 (Pediatric Use).}
  \label{fig:calibration_FDA}
\end{figure}

\paragraph{Calibration Analysis.}
Calibration curves (Figure~\ref{fig:calibration_FDA}) reveal that GPT-4.1 is \textit{underconfident} on FDA labels: predicted confidence $\hat{p}$ consistently falls below empirical accuracy. Expected Calibration Error (ECE) ranges from 0.012 (Adverse Reactions, $n{=}63{,}442$) to 0.214 (Pediatric Use, $n{=}1{,}954$). The underconfidence pattern is consistent across most sections, with the model assigning conservative probabilities to entities it extracts correctly. Pediatric Use is a notable exception: its calibration curve falls below the diagonal, indicating overconfidence, with the highest ECE among all sections. This contrasts with the overconfidence typically observed in free-text generation tasks~\cite{geng2024survey} and suggests that the structured, formulaic nature of FDA labels makes extraction easier than the model's confidence suggests for most sections.

\paragraph{FDR Sweep Analysis.}
Table~\ref{tab:fda-sweep} presents a sweep across $\alpha$ values using a single global threshold. The global baseline FDR across all 110,664 entities is approximately 2.3\%. As a result, any $\alpha \geq 0.03$ accepts all extractions, since the baseline error rate already satisfies the FDR target. At $\alpha = 0.01$, no threshold can bring FDR below the target, resulting in full rejection.

\begin{table}[H]
\centering
\caption{FDR-controlled conformal prediction sweep across $\alpha$ for FDA drug label extraction (GPT-4.1, $n{=}110{,}664$ entities, global threshold). At $\alpha \geq 0.03$ the global baseline FDR (2.3\%) is already below target, so nearly all entities are accepted; per-section thresholds (Table~\ref{tab:fda-per-section}) reveal the heterogeneity masked by global pooling.}
\label{tab:fda-sweep}
\begin{tabular}{cccccc}
\toprule
$\alpha$ & $\tau_c$ & $\hat{p}_{\min}$ & Rej.\% & Cov. & Prec. \\
\midrule
0.01 & $\infty$ & -- & 100.0 & 0.000 & -- \\
0.03 & $-0.49$ & .380 &  0.0 & 1.000 & 0.977 \\
0.05 & $-0.49$ & .380 &  0.0 & 1.000 & 0.977 \\
0.10 & $-0.49$ & .380 &  0.0 & 1.000 & 0.977 \\
0.15 & $-0.49$ & .380 &  0.0 & 1.000 & 0.977 \\
0.20 & $-0.49$ & .380 &  0.0 & 1.000 & 0.977 \\
\bottomrule
\end{tabular}
\end{table}

\paragraph{Per-Section FDR Results.}
The global sweep masks substantial per-section heterogeneity (Table~\ref{tab:fda-per-section}). At $\alpha = 0.05$, high-accuracy sections such as Adverse Reactions and Indications \& Usage accept all entities with 0\% rejection, while sections with higher baseline error rates require aggressive filtering: Drug Interactions rejects 59.8\% of extractions, Contraindications rejects 41.5\%, and Pediatric Use rejects 100\%, meaning that no threshold can bring its FDR below 0.05. Pediatric Use is the clear outlier with baseline accuracy of only 74.3\% and the highest ECE (0.214); unlike most other sections where the model is underconfident, Pediatric Use exhibits overconfidence, and FDR control correctly identifies it as unsuitable for automated acceptance at this error tolerance.

\begin{table}[H]
\centering
\caption{Calibration metrics and FDR-controlled conformal prediction for FDA drug label extraction with GPT-4.1 ($\alpha = 0.05$, $n{=}1{,}000$ drugs). $\tau_c$ is the per-section FDR threshold in log-odds; $\hat{p}_{\min} = \sigma(\tau_c)$ is the equivalent confidence cutoff. FDR control guarantees precision $\geq 1{-}\alpha$ among accepted entities. Warnings section omitted (no valid entities).}
\label{tab:fda-per-section}
\resizebox{\textwidth}{!}{%
\begin{tabular}{lccccccccc}
\toprule
Section & \# Entities & Brier$\downarrow$ & ECE$\downarrow$ & FactS. & $\tau_c$ & $\hat{p}_{\min}$ & Rej.\% & Cov. & Prec. \\
\midrule
Adverse Reactions      & 63{,}442 & 0.013 & 0.012 & 2.98 & $-0.09$ & .479 &  0.0 & 1.000 & 0.987 \\
Indications \& Usage   & 34{,}895 & 0.022 & 0.018 & 2.94 & $0.85$ & .701 &  0.0 & 1.000 & 0.979 \\
Drug Interactions      &  7{,}251 & 0.051 & 0.055 & 2.92 & $10.52$ & ${>}.999$ & 59.8 & 0.400 & 0.948 \\
Contraindications      &  2{,}109 & 0.048 & 0.041 & 2.90 & $7.65$ & ${>}.999$ & 41.5 & 0.590 & 0.966 \\
Pregnancy              &      545 & 0.005 & 0.004 & 2.99 & $2.71$ & .938 &  0.4 & 0.996 & 0.996 \\
Pediatric Use          &  1{,}954 & 0.236 & 0.214 & 2.52 & $\infty$ & -- & 100.0 & 0.000 & -- \\
Geriatric Use          &      468 & 0.000 & 0.007 & 3.00 & $3.06$ & .955 &  0.0 & 1.000 & 1.000 \\
\midrule
All sections           & 110{,}664 & 0.023 & 0.018 & 2.95 & $-0.49$ & .380 &  0.0 & 1.000 & 0.977 \\
\bottomrule
\end{tabular}%
}
\end{table}

\subsection{RadGraph Entity Extraction}

We evaluated extraction on 100 MIMIC-CXR test reports against RadGraph gold annotations using both GPT-4.1 and Llama-4-Maverick.

\paragraph{Extraction Quality.}
With 5-shot prompting, GPT-4.1 achieves entity F1 of 0.826 (precision 0.798, recall 0.856) and relation F1 of 0.502. Llama-4-Maverick achieves entity F1 of 0.841 (precision 0.865, recall 0.818) and relation F1 of 0.490 (Table~\ref{tab:radgraph-evaluation}). The two models exhibit complementary strengths: GPT-4.1 has higher recall while Llama-4-Maverick has higher precision. Relation extraction remains challenging for both models at around F1 = 0.50. Zero-shot performance degrades substantially for both models, with entity F1 dropping by 8.0 percentage points for GPT-4.1 and 11.7 points for Llama-4-Maverick, and relation F1 dropping by over 22 points for both, indicating that the RadGraph schema's labeling conventions require in-context demonstration.

\begin{table}[H]
\centering
\caption{RadGraph entity and relation extraction on 100 test reports. Few-shot prompting (5 examples from dev set) substantially improves both entity and relation extraction compared to zero-shot.}
\label{tab:radgraph-evaluation}
\begin{tabular}{llcccccc}
\toprule
& & \multicolumn{3}{c}{Entity} & \multicolumn{3}{c}{Relation} \\
\cmidrule(lr){3-5} \cmidrule(lr){6-8}
Model & Setting & Prec. & Rec. & F1 & Prec. & Rec. & F1 \\
\midrule
GPT-4.1 & 5-shot & .798 & .856 & .826 & .484 & .521 & .502 \\
GPT-4.1 & 0-shot & .787 & .710 & .746 & .301 & .252 & .274 \\
Llama-4-Maverick & 5-shot & .865 & .818 & .841 & .526 & .458 & .490 \\
Llama-4-Maverick & 0-shot & .783 & .674 & .724 & .228 & .227 & .228 \\
\bottomrule
\end{tabular}
\end{table}

\begin{figure}[htbp]
  \centering
  \includegraphics[width=1\textwidth]{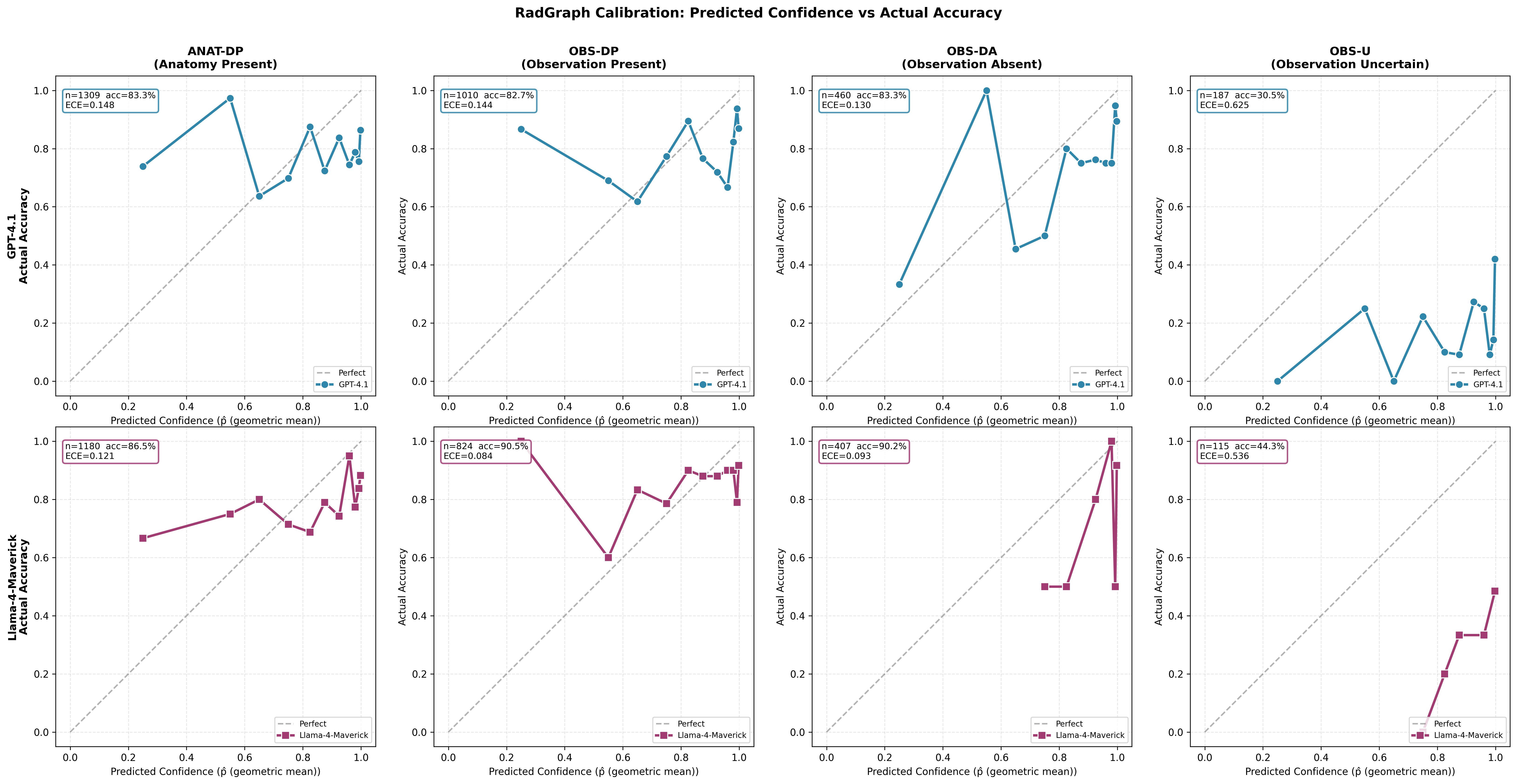}
  \caption{Per-label calibration curves for GPT-4.1 and Llama-4-Maverick (5-shot, 100 CXR reports~\cite{johnson2016mimic}). Predicted span confidence $\hat{p}$ versus empirical accuracy across four entity categories. Both models are overconfident (curves below diagonal). Llama-4-Maverick achieves lower ECE overall (0.085 vs.\ 0.147), while OBS-U remains poorly calibrated for both models.}
  \label{fig:calibration_RadGraph}
\end{figure}

\paragraph{Calibration Analysis.}
In stark contrast to FDA labels, both models are \textit{overconfident} on radiology reports (Figure~\ref{fig:calibration_RadGraph}): they assign near-certain probabilities to incorrect extractions. Llama-4-Maverick achieves lower ECE than GPT-4.1 across most categories (overall 0.085 vs.\ 0.147), indicating better-calibrated confidence despite similar extraction quality. OBS-U (uncertain observations) remains the most challenging category for both models, with ECE exceeding 0.40, reflecting the inherent ambiguity of hedging language in radiology reports (e.g., ``cannot exclude,'' ``possibly representing'').

\paragraph{FDR Sweep Analysis.}
Table~\ref{tab:radgraph-sweep} presents a sweep across $\alpha$ values using global thresholds. At $\alpha = 0.05$, both models reject all extractions: the baseline error rate of 15--20\% makes it impossible to guarantee that fewer than 5\% of accepted extractions are incorrect. A sharp transition occurs at $\alpha = 0.10$, where Llama-4-Maverick accepts 80.4\% of extractions (19.6\% rejection) while GPT-4.1 accepts only 40.7\% (59.3\% rejection). This difference reflects Llama-4-Maverick's better calibration: its confidence scores separate correct from incorrect extractions more effectively, allowing the FDR threshold to be satisfied with less aggressive filtering. At $\alpha \geq 0.15$, Llama-4-Maverick accepts nearly all extractions while GPT-4.1 continues to require substantial rejection until $\alpha = 0.25$.

\begin{table}[H]
\centering
\caption{FDR-controlled conformal prediction sweep across $\alpha$ for RadGraph entity extraction (global threshold, 5-shot). At $\alpha < 0.10$, neither model can guarantee the target precision, resulting in full rejection. The sharp transition at $\alpha = 0.10$ reflects the baseline error rate (${\sim}15{-}20\%$); Llama-4-Maverick's better calibration yields a lower threshold and less rejection at the same $\alpha$.}
\label{tab:radgraph-sweep}
\begin{tabular}{lcccccc}
\toprule
& \multicolumn{3}{c}{GPT-4.1} & \multicolumn{3}{c}{Llama-4-Maverick} \\
\cmidrule(lr){2-4} \cmidrule(lr){5-7}
$\alpha$ & Rej.\% & Cov. & Prec. & Rej.\% & Cov. & Prec. \\
\midrule
0.05 & 100.0 & 0.000 & -- & 100.0 & 0.000 & -- \\
0.10 &  59.3 & 0.445 & 0.876 &  19.6 & 0.824 & 0.870 \\
0.15 &  41.7 & 0.617 & 0.849 &   0.2 & 0.998 & 0.849 \\
0.20 &   6.8 & 0.936 & 0.805 &   0.2 & 0.998 & 0.849 \\
0.25 &   0.1 & 0.999 & 0.802 &   0.2 & 0.998 & 0.849 \\
\bottomrule
\end{tabular}
\end{table}

\paragraph{Per-Category FDR Results.}
Per-category results at $\alpha = 0.10$ (Table~\ref{tab:radgraph-per-label}) reveal further heterogeneity. Both models reject 100\% of OBS-U extractions, confirming that uncertain observations cannot be reliably extracted by either model at this error tolerance. For the remaining categories, the two models exhibit strikingly different patterns. Llama-4-Maverick accepts nearly all OBS-DP and OBS-DA extractions (0.2\% and 0.0\% rejection), while GPT-4.1 rejects 52.1\% and 62.6\% of those same categories. For ANAT-DP, both models require strict thresholds ($\hat{p}_{\min} > 0.999$), but GPT-4.1 rejects a larger fraction (57.3\% vs.\ 41.4\%). These cross-model differences arise despite similar extraction quality and demonstrate that FDR-controlling thresholds depend not only on accuracy but on how well a model's confidence scores discriminate between correct and incorrect extractions.

\begin{table}[H]
\centering
\caption{Per-category entity extraction and FDR-controlled conformal prediction on 100 RadGraph test reports ($\alpha = 0.10$, 5-shot). $\tau_c$ is the per-category FDR threshold in log-odds; $\hat{p}_{\min} = \sigma(\tau_c)$ is the equivalent confidence cutoff. FDR control guarantees precision $\geq 1{-}\alpha$ among accepted entities.}
\label{tab:radgraph-per-label}
\resizebox{\textwidth}{!}{%
\begin{tabular}{lcccccccccccccc}
\toprule
& \multicolumn{7}{c}{GPT-4.1} & \multicolumn{7}{c}{Llama-4-Maverick} \\
\cmidrule(lr){2-8} \cmidrule(lr){9-15}
Category & Prec. & Rec. & F1 & ECE$\downarrow$ & $\tau_c$ & $\hat{p}_{\min}$ & Rej.\% & Prec. & Rec. & F1 & ECE$\downarrow$ & $\tau_c$ & $\hat{p}_{\min}$ & Rej.\% \\
\midrule
\textsc{anat-dp} & .834 & .933 & .880 & .128 & $9.92$ & ${>}.999$ & 57.3 & .865 & .893 & .879 & .085 & $11.11$ & ${>}.999$ & 41.4 \\
\textsc{obs-dp}  & .827 & .769 & .797 & .140 & $6.45$ & .998 & 52.1 & .905 & .724 & .804 & .064 & $0.00$ & .500 &  0.2 \\
\textsc{obs-da}  & .833 & .914 & .871 & .102 & $13.12$ & ${>}.999$ & 62.6 & .902 & .900 & .901 & .051 & $0.00$ & .500 &  0.0 \\
\textsc{obs-u}   & .305 & .648 & .415 & .525 & $\infty$ & -- & 100.0 & .444 & .580 & .502 & .396 & $\infty$ & -- & 100.0 \\
\bottomrule
\end{tabular}%
}
\end{table}

\subsection{Cross-Domain Calibration Reversal}

Our most striking finding is the reversal of miscalibration direction across domains. On FDA labels, GPT-4.1 is systematically underconfident across most sections (ECE 0.004--0.055, curves above the diagonal in Figure~\ref{fig:calibration_FDA}), with Pediatric Use as the exception showing overconfidence (ECE 0.214). On radiology reports, the same model family is overconfident (ECE 0.102--0.525, curves below the diagonal in Figure~\ref{fig:calibration_RadGraph}).

We attribute this reversal to differences in document structure. FDA labels follow standardized formatting with regulatory language, making entity boundaries clear and extraction straightforward. The model assigns conservative probabilities, perhaps because the structured format resembles training data where careful hedging is rewarded. Radiology reports, by contrast, use terse, variable-format clinical shorthand with implicit negation and hedging, making extraction genuinely harder. The model assigns high confidence to entities that appear linguistically plausible but are semantically incorrect (e.g., labeling a finding as ``definitely present'' when the report says ``cannot be excluded'').

The consequences for FDR control are dramatic. On FDA labels, the global baseline FDR of 2.3\% trivially satisfies $\alpha = 0.05$, though per-section analysis reveals three sections requiring 41--100\% rejection. On RadGraph, neither model can satisfy $\alpha = 0.05$ at all, and even at $\alpha = 0.10$ the two models require very different levels of filtering (19.6\% vs.\ 59.3\% global rejection). This finding has important implications for clinical deployment: a single calibration strategy cannot work across document types, and FDR control exposes heterogeneity that coverage-based approaches and global thresholds obscure.

\section{Conclusion}

We have presented a conformal prediction framework for LLM-based medical entity extraction that provides finite-sample FDR guarantees across two clinical domains with fundamentally different calibration characteristics. Our experiments demonstrate that: (1) LLM calibration direction reverses between structured FDA labels (underconfident) and free-text radiology reports (overconfident); (2) FDR-controlling conformal prediction adapts automatically to both regimes, with per-category thresholds revealing heterogeneity that global pooling masks; (3) sweep analysis across $\alpha$ values exposes sharp transitions in acceptance behavior tied to the baseline error structure of each domain; (4) cross-model comparison shows that FDR-controlling thresholds depend on confidence discriminability, not just extraction accuracy, with Llama-4-Maverick requiring far less rejection than GPT-4.1 at the same $\alpha$ on RadGraph despite similar F1.

\paragraph{Limitations.}
Our framework requires access to token-level log-probabilities, which some frontier models do not currently expose. The FDA label verification relies on an LLM-as-a-judge (GPT-5-mini), which may introduce systematic biases in the ground truth. The RadGraph evaluation is limited to 100 test reports, which constrains the statistical power of per-category conformal analysis, particularly for rare categories like OBS-U ($n = 88$ gold entities). Additionally, our approach assumes exchangeability within each domain, which may be violated if certain drug types or report formats are systematically harder.

\paragraph{Future Work.}
Several directions remain open. First, extending conformal guarantees to black-box models without log-probabilities via surrogate confidence estimation or verbalized uncertainty. Second, comparison with post-hoc calibration baselines (temperature scaling, Platt scaling) to quantify the advantage of conformal prediction's formal guarantees. Third, deploying the framework in a clinical workflow study to measure the impact of conformal filtering on downstream clinical decision-making. Fourth, exploring adaptive conformal prediction methods that can handle distribution shift across time as drug labels and reporting practices evolve.

\footnotetext{Declaration of AI Assistance: The authors used GitHub Copilot for writing or drafting manuscript content, and refinement or formatting of code reported in the submitted manuscript. After using this technology, the authors reviewed the results and take full responsibility for the contents of the manuscript.}

\bibliographystyle{splncs04}
\bibliography{biblio}

\end{document}